\documentclass[10pt,twocolumn,letterpaper]{article}

\usepackage[margin=1cm]{geometry}
\usepackage{tikz}
\usepackage{pgf-pie}
\usepackage{stfloats}
\usepackage{pifont}
\usepackage{tcolorbox}
\usepackage{caption, subcaption}
\usepackage[accsupp]{axessibility} 

\usepackage[pagenumbers]{wacv} 

\tcbset{
    colback=gray!10,   
    colframe=gray!40,  
    arc=4pt,           
    boxrule=0.5pt,
    left=6pt,right=6pt,top=6pt,bottom=6pt
}

\definecolor{wacvblue}{rgb}{0.21,0.49,0.74}
\usepackage[pagebackref,breaklinks,colorlinks,allcolors=wacvblue]{hyperref}

\def\wacvPaperID{11}
\def\confName{WACV}
\def\confYear{2026}

\title{SOVABench: A Vehicle Surveillance Action Retrieval Benchmark for Multimodal Large Language Models}

\author{Oriol Rabasseda$^{2}$\\
{\tt\small oriol.rabasseda@gmail.com}
\and
Zenjie Li$^{1}$\\
{\tt\small zli@milestone.dk}
\and
Kamal Nasrollahi$^{1,3}$\\
{\tt\small kna@milestone.dk}
\and
Sergio Escalera$^{2,3}$\\
{\tt\small sescalera@ub.edu}\\
\and
$^{1}$Milestone Systems A/S\\
{\small Banemarksvej 50, Brøndby}
\and
$^{2}$Universitat de Barcelona and \\
Computer Vision Center\\
{\small Gran Via de les Corts Catalanes 585, Barcelona} \\ {\small Campus UAB, Edifici O, Cerdanyola del Vallès}
\and
$^{3}$Aalborg Universitet\\
{\small Fredrik Bajers Vej 7K, Aalborg Øst}
}

\begin{document}
\AddToShipoutPicture*{%
  \AtTextLowerLeft{%
    \put(0,\LenToUnit{-12pt}){%
      \parbox{\textwidth}{\centering\footnotesize\itshape This paper was published in the proceedings of the IEEE/CVF Winter Conference on Applications of Computer Vision (WACV) Workshops, 2026.}%
    }%
  }%
}
\maketitle
\begin{abstract}
Automatic identification of events and recurrent behavior analysis are critical for video surveillance. However, most existing content-based video retrieval benchmarks focus on scene-level similarity and do not evaluate the action discrimination required in surveillance. To address this gap, we introduce SOVABench (Surveillance Opposite Vehicle Actions Benchmark), a real-world retrieval benchmark built from surveillance footage and centered on vehicle-related actions. SOVABench defines two evaluation protocols (inter-pair and intra-pair) to assess cross-action discrimination and temporal direction understanding. Although action distinctions are generally intuitive for human observers, our experiments show that they remain challenging for state-of-the-art vision and multimodal models.

Leveraging the visual reasoning and instruction-following capabilities of Multimodal Large Language Models (MLLMs), we present a training-free framework for producing interpretable embeddings from MLLM-generated descriptions for both images and videos. The framework achieves strong performance on SOVABench as well as on several spatial and counting benchmarks where contrastive Vision-Language Models often fail. The code, annotations, and instructions to construct the benchmark are publicly available\footnote{\texttt{https://github.com/oriol-rabasseda/sovabench.git}}.
\end{abstract}  
\section{Introduction} \label{sec:intro}
\begin{figure}[ht]
    \centering
    \begin{subfigure}{\linewidth}
        \centering
        \begin{subfigure}[t]{\textwidth}
            \centering
            \includegraphics[width=\linewidth]{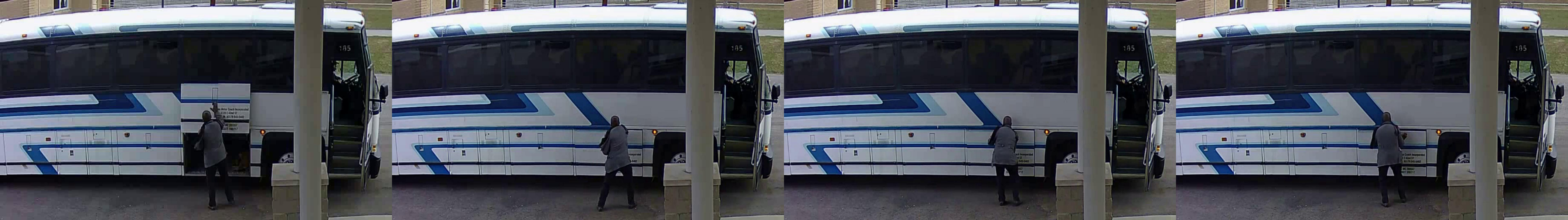}
        \end{subfigure} \\
        \vspace{3pt}
        \begin{subfigure}[t]{\textwidth}
            \centering
            \includegraphics[width=\linewidth]{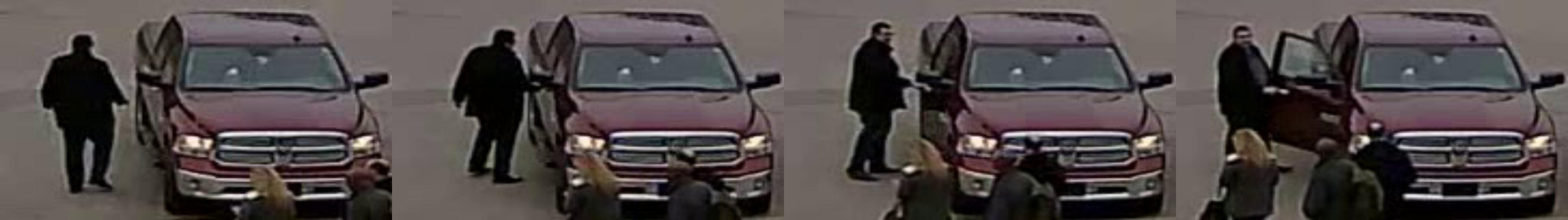}
        \end{subfigure} \\
        \vspace{3pt}
        \begin{subfigure}[t]{\textwidth}
            \centering
            \includegraphics[width=\linewidth]{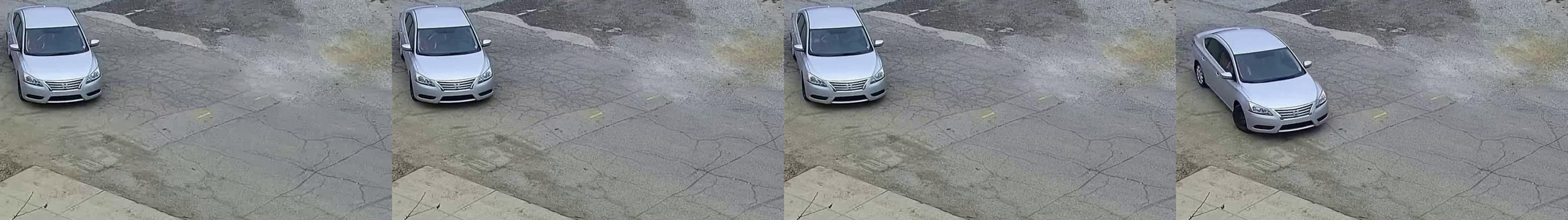}
        \end{subfigure} \\
        \vspace{3pt}
        \begin{subfigure}[t]{\textwidth}
            \centering
            \includegraphics[width=\linewidth]{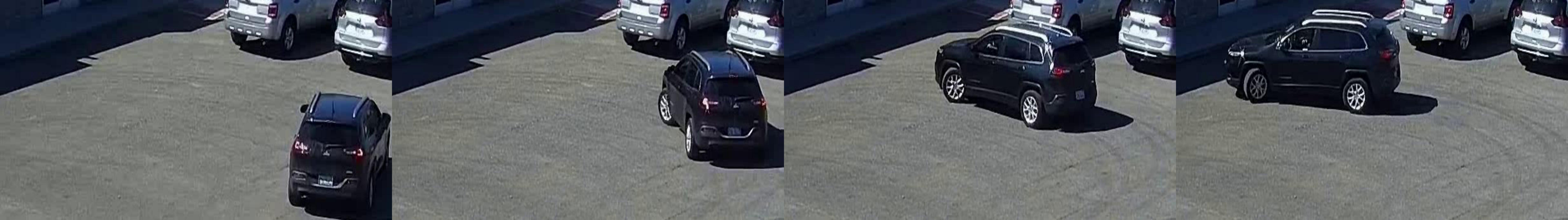}
        \end{subfigure}
        \caption{}
        \label{fig:examples}
    \end{subfigure}
    \begin{subfigure}{\linewidth}
        \centering
        \includegraphics[width=\linewidth]{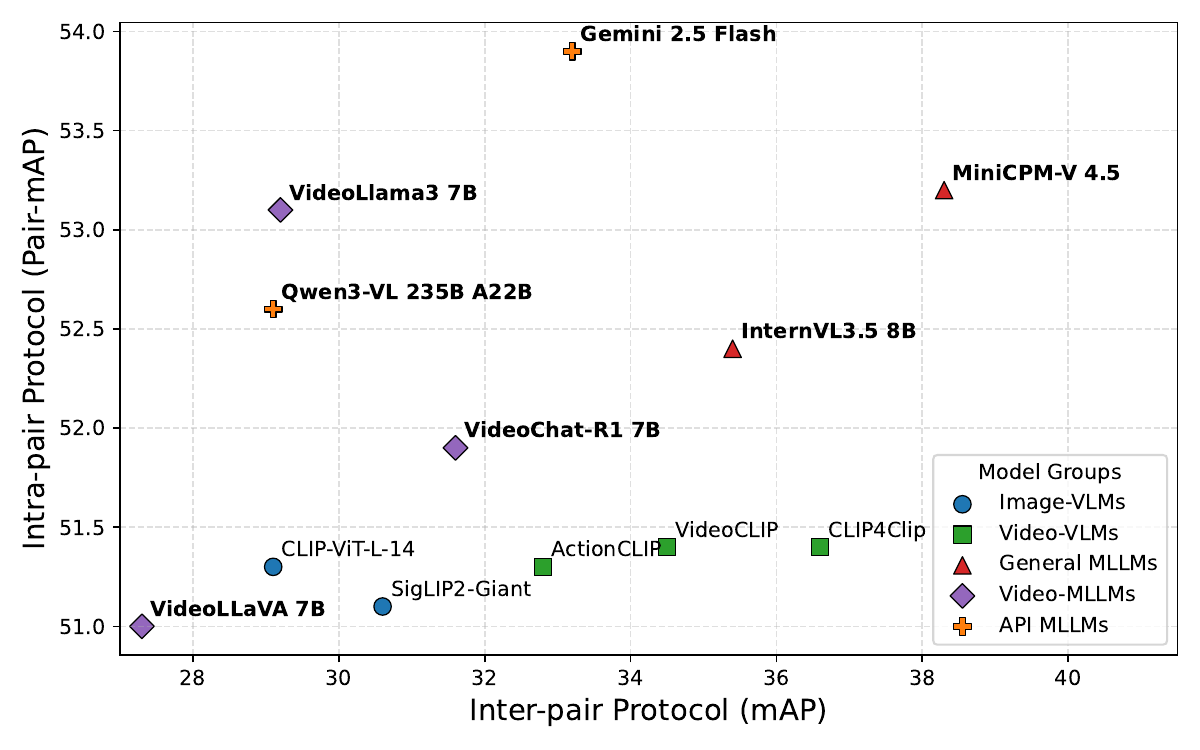}
        \caption{}
        \label{fig:plot}
    \end{subfigure}
    \vspace{-17pt}
    \caption{\textbf{Samples and performance in SOVABench.} (a) Illustrative samples of the constructed benchmark of four different actions (close trunk, open vehicle door, start, and turn left), and (b) comparison of methods for the two evaluation protocols in SOVABench. Methods include MLLMs using the MLLM-to-Embedding framework to obtaining embeddings (bold) and contrastive VLMs. For reference, random values are 3.4 mAP and 50.3 Pair-mAP in Inter-pair and Intra-pair protocols respectively.}
    \vspace{-15pt}
\end{figure}

In the video surveillance domain, a challenging task is to automatically identify similar events. This capability is essential in applications such as alarm filtering and recurrent event detection. To tackle this task, the required video retrieval system should capture not-only high-level semantics but also information about object relations, motion patterns, and temporal dynamics. However, existing content-based video retrieval (CBVR) benchmarks usually evaluate scene similarity without focusing on action recognition in the surveillance domain \cite{revaud2013event, kordopatis2019fivr, jiang2014vcdb}.

To address the lack of a dedicated, real-world dataset for evaluating this capability, we introduce \textbf{SOVABench} (Surveillance Opposite Vehicle Actions Benchmark). SOVABench reorganizes and labels vehicle surveillance footage into a retrieval-oriented benchmark built around opposite action pairs (\eg, loading vs. unloading a vehicle). The benchmark defines two complementary evaluation protocols:
(1) inter-pair retrieval, which assesses a model’s ability to discriminate between different pairs, and
(2) intra-pair retrieval, a more challenging setting that requires distinguishing opposite actions differing primarily in their temporal evolution. Using the structure of opposite action pairs, the two protocols jointly provide a systematic evaluation of how effectively embeddings represent action semantics and temporal direction. Examples of video clips from SOVABench are shown in Figure~\ref{fig:examples}.

Alongside the benchmark, we explore a simple, training-free, instruction-following embedding framework that uses Multimodal Large Language Models (MLLMs) as black-box unified visual encoders for both images and videos. Recent progress in instruction following \cite{liu2023visual} and visual reasoning \cite{alayrac2022flamingo} suggests that textual explanations generated by MLLMs may provide more task-sensitive representations than global embeddings from contrastive Vision-Language Models (VLMs). Our framework extracts these explanations and converts them into sentence-level embeddings for comparison using a maximum similarity operation.

Although surveillance is inherently video-focused, we first validate the ability of the framework to take images on image-based tasks that capture aspects of scene understanding relevant to surveillance (spatial relationships and object counting). The results show that the framework outperforms CLIP \cite{radford2021learning} in these tasks. We then apply the framework to the SOVABench retrieval settings, showing that its instruction-conditioned embeddings are effective for action discrimination in real-world surveillance videos (see Figure \ref{fig:plot}). Our contributions can be summarized as follows.

\begin{itemize}
    \item[(1)] We introduce SOVABench, the first content-based video retrieval benchmark built from real-world vehicle-surveillance footage, designed to evaluate action discrimination and temporal direction understanding.
    \item[(2)] We present a simple, training-free, instruction-following embedding framework that uses MLLMs as visual encoders, obtaining strong results on two image-based visual tasks and on SOVABench.
\end{itemize}

The remainder of this paper is organized as follows. Section~\ref{sec:rel_work} reviews related work on embedding-based and surveillance benchmarks, visual embeddings, and instruction-following MLLMs. Section~\ref{sec:benchmark} introduces SOVABench and its evaluation protocols. Section~\ref{sec:method} details the embedding framework used. Section~\ref{sec:experiments} presents experiments and results, and Section~\ref{sec:conclusions} concludes the paper and outlines future directions.
\section{Related work} \label{sec:rel_work}
\subsection{Benchmarks for Multimodal and Video Surveillance}
\paragraph{Multimodal Benchmarks.}
Generalist multimodal reasoning benchmarks \cite{tong2024eyes, thrush2022winoground, parcalabescu2021valse, hsieh2023sugarcrepe} evaluate image-text compositional alignment, while specialized datasets target tasks such as spatial reasoning \cite{liu2023vsr, wang2025spatialclip}, numerical reasoning \cite{paiss2023teaching}, and logical inference \cite{zhou2025logic}. Although useful for validating our framework, these datasets tackle classification and revolve around static images that do not model temporal progression and retrieval unlike SOVABench.

\vspace{-6pt}
\paragraph{Content-Based Video Retrieval Benchmarks.}
CBVR benchmarks assess how well visual embeddings retrieve semantically similar videos for tasks such as clip repetition retrieval \cite{jiang2014vcdb}, incident retrieval \cite{kordopatis2019fivr}, and event retrieval \cite{revaud2013event}.  Within this category, SOVABench focuses on action retrieval in vehicle-centric surveillance scenarios.

\vspace{-6pt}
\paragraph{Video Surveillance Benchmarks.}
Video surveillance datasets such as MEVA \cite{corona2021meva} and VIRAT \cite{oh2011large} provide annotated footage of human and vehicle activities. More recently, SurveillanceVQA-589K \cite{liu2025surveillancevqa} was released to evaluate the capabilities of MLLMs in the surveillance domain. However, these datasets do not support CBVR.


\subsection{Multimodal Large Language Models}
Recent MLLMs such as Qwen3-VL \cite{bai2025qwen2, yang2025qwen3}, MiniCPM-V 4.5 \cite{yu2025minicpm}, and InternVL-3.5 \cite{wang2025internvl3} show strong visual reasoning and instruction-following capabilities. Unlike contrastive VLMs that produce fixed global embeddings, MLLMs condition outputs on user instructions \cite{liu2023visual, dai2023instructblip}, often yielding richer and more interpretable responses. Video-centric variants \cite{zhang2025videollama, zhang2024video, li2025videochat} focus on temporal modeling. In this work, we use MLLMs as black-box describers, deriving task-aware embeddings from their generated text.

\subsection{Visual Embedding Learning}
\paragraph{Contrastive Vision–Language Models.}
Contrastive VLMs such as CLIP \cite{radford2021learning} and SigLIP2 \cite{tschannen2025siglip} provide strong general-purpose image embeddings, but their task-agnostic representations often struggle with fine-grained reasoning \cite{li2024exploring}. Although not addressing this limitation, extensions like ICE \cite{yu2025image} have explored the incorporation of textual captions at inference time  to improve classification. However, ICE remains tied to captioning models rather than instruction-following MLLMs. Video–text alignment models \cite{xu2021videoclip, luo2022clip4clip, wang2023actionclip} extend contrastive approaches to the temporal dimension but inherit the same limitations in reasoning granularity.

\vspace{-6pt}
\paragraph{Instruction- and Task-Aware Embeddings.}
Instruction-tuned text encoders \cite{su2022one, asai2022task} and their multimodal counterparts \cite{cui2025think, jiang2024vlm2vec} show that conditioning visual representations on textual instructions improves generalization. However, existing multimodal approaches typically require new training and pre-defined instruction sets. In contrast, our framework is training-free and relies solely on the open-ended outputs generated by off-the-shelf MLLMs.

\vspace{-6pt}
\paragraph{Embeddings from MLLMs' Outputs.}
Works such as CoLLM \cite{huynh2025collm}, Think-And-Embed \cite{cui2025think}, and Shih et al. \cite{shih2024visual} explore image embedding generation or enrichment through MLLM outputs. We follow this direction, and extend it with a sentence-level embedding strategy that accommodates arbitrarily long MLLM descriptions.
\section{SOVABench} \label{sec:benchmark}
We introduce \textbf{SOVABench} (Surveillance Opposite Vehicle Actions Benchmark), a surveillance benchmark designed to evaluate CBVR in vehicle surveillance scenarios. Existing CBVR benchmarks do not target action retrieval, which requires understanding motion differences rather than scene similarity (see Table~\ref{tab:retrieval_datasets}).

A central feature of SOVABench is its focus on opposite vehicle actions (\eg, loading vs. unloading), which are visually and semantically similar but differ in temporal evolution. This proximity makes them suitable for probing whether embeddings capture temporal cues, while also enabling a coarser granularity where each opposite-action pair can be merged into a unified class to assess broader action discrimination. To accommodate these complementary levels of granularity, SOVABench defines two evaluation protocols: inter-pair and intra-pair. Together, these protocols provide a systematic analysis of when embeddings succeed in representing actions and their temporal progression.

SOVABench is constructed from two surveillance datasets: MEVA \cite{corona2021meva} and the VIRAT validation set \cite{oh2011large}. From these sources, we extract vehicle-related activities and organize them into actions that constitute the set of queries. Action classes are structured into pairs of opposite actions, as listed in Table~\ref{tab:opposing_actions}. The footage comes from different scenes and backgrounds, making it not possible to reliably infer the action class based only on context. To further focus on the relevant actions and suppress background information, we identify the participating objects for each activity and define a spatial region of interest enclosing all actors throughout the activity. Because vehicle-surveillance cameras are typically static, this results in stable video crops that isolate the action while maintaining the relevant temporal dynamics. In addition, each clip is temporally aligned with the duration of its annotated activity, ensuring that every clip captures a single action.

\begin{table}[ht]
    \centering
    \begin{tabular}{@{} l l | c c @{}}
        \toprule
        \textbf{Action} & \textbf{Opposite} & \textbf{Inter-} & \textbf{Intra-} \\
         & \textbf{Action} & \textbf{pair} & \textbf{pair} \\
        \midrule
        Drive forward & Reverse & \ding{55} & \ding{52} \\
        Enter vehicle & Exit vehicle & \ding{52} & \ding{52} \\
        Load vehicle & Unload vehicle & \ding{52} & \ding{52} \\
        Open trunk & Close trunk & \ding{52} & \ding{52} \\
        Open vehicle door & Close vehicle door & \ding{52} & \ding{52} \\
        Start & Stop & \ding{52} & \ding{52} \\
        Turn left & Turn right & \ding{52} & \ding{52}
    \end{tabular}
    \caption{\textbf{Opposite vehicle-related actions included in SOVABench.} Check marks indicate the use of the pair for each evaluation protocol.}
    \label{tab:opposing_actions}
    \vspace{-4pt}
\end{table}

\subsection{Inter-pair Evaluation Protocol}
This protocol evaluates the model’s ability to distinguish between different pairs. Using the semantic similarity of each pair of opposite actions, we treat them as unified classes. Retrieval evaluation follows a one-versus-all setup using sample-level mean Average Precision (mAP), the most used metric in CBVR benchmarks given its suitability for ranked retrieval. The six resulting action-pair query classes are listed in Table~\ref{tab:opposing_actions}. We exclude the pair $<$Drive forward, Reverse$>$, as these motions always co-occur with other vehicle-movement actions (\eg $<$Turn left, Turn right$>$), preventing them from forming an isolated class.

Moreover, we also include human-only surveillance activities from the same source datasets as distracting samples. These samples are relevant to the surveillance domain but do not involve vehicles, which poses an additional challenge while being semantically separate from the set of vehicle-action queries. The effect of the incorporation of these distracting samples to the benchmark is evaluated in the supplementary material.

All query samples exhibit non-overlapping actions, with each video clip containing exactly one class and no other visible classes. Temporal action boundaries and the absence of overlapping actions were reviewed by human annotators, ensuring that events are temporally complete and that visual evidence of the action is observable. These properties guarantee that the dataset is clean, well-structured, and suitable for evaluating discrimination capabilities in retrieval.

\begin{figure*}[ht]
    \centering
    \begin{minipage}[b]{0.27\textwidth}
        \centering
        \includegraphics[width=\linewidth]{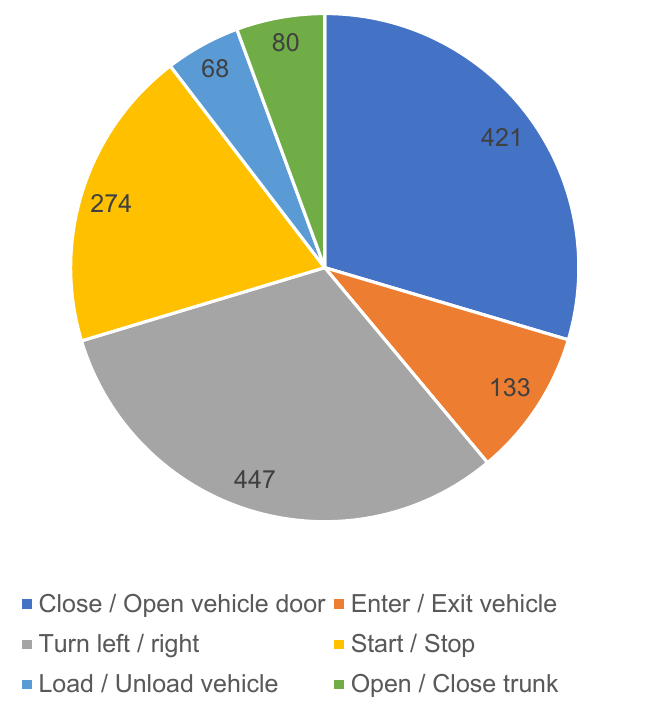}
        \subcaption{}
    \end{minipage}
    \begin{minipage}[b]{0.36\textwidth}
        \centering
        \includegraphics[width=\linewidth]{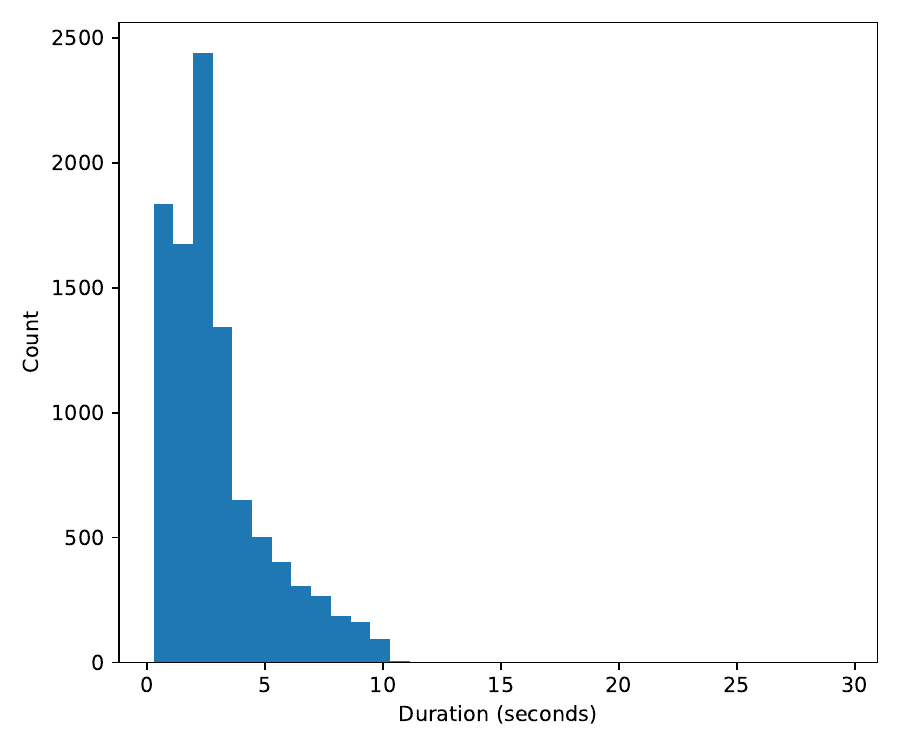}
        \subcaption{}
    \end{minipage}
    \begin{minipage}[b]{0.36\textwidth}
        \centering
        \includegraphics[width=\linewidth]{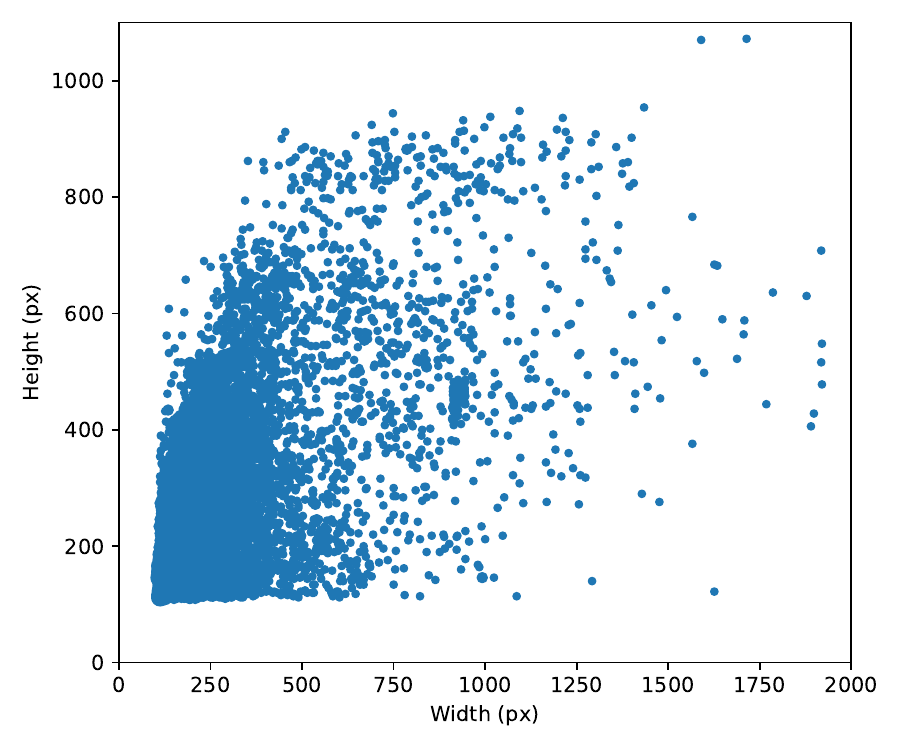}
        \subcaption{}
    \end{minipage}
    \caption{
        \textbf{Statistical overview of the SOVABench (Inter-pair) benchmark.} The class count of the SOVABench (Intra-pair) benchmark can be accessed in the supplementary material.
        (a) Sample count per class in queries.
        (b) Distribution of clip durations.
        (c) Resolution distribution resulting from spatial cropping.
    }
    \label{fig:dataset_stats}
\end{figure*}

The resulting SOVABench (Inter-pair) dataset contains 1,423 queries and a total of 9,882 samples. Figure~\ref{fig:dataset_stats} shows the number of video samples per class of the query set and the video duration statistics of the entire dataset (queries + distracting samples), with clips that typically range from 1 to 10 seconds. In addition, spatial cropping produces videos with non-usual frame shapes, capturing the diversity of object scales in surveillance and posing a challenge for models. All samples are used for testing, reflecting the open-world nature of surveillance.

Unlike previous CBVR benchmarks (Table~\ref{tab:retrieval_datasets}), SOVABench is specifically designed for video-based action retrieval. Although the total number of samples is smaller than in existing benchmarks, it provides a large number of queries. Only these queries are used for evaluation, while the remaining distracting samples increase difficulty and are typically less curated.

\begin{table*}[ht]
    \centering
    \begin{tabular}{l c c c c l}
        \toprule
        \textbf{Dataset} & \textbf{Data} & \textbf{Metric} & \textbf{\#Samples} & \textbf{\#Queries} & \textbf{Retrieval task} \\
        \midrule
        FIVR-200K \cite{kordopatis2019fivr} & Video & mAP & 226k & 100 & Fine-grained Incident Retrieval \\
        VCDB \cite{jiang2014vcdb} & Video & Prec. \& Recall & 100.5k & 528 & Near-Duplicate Retrieval \\
        EVVE \cite{revaud2013event} & Video & mAP & 102.4k & 620 & Event Retrieval \\
        \midrule
        \textbf{SOVABench (Inter-pair)} & Video & mAP & 9.9k & 1,423 & Action Retrieval \\
        \textbf{SOVABench (Intra-pair)} & Video & Pair-mAP & 2.3k & 2,300 & Opposite Action Retrieval \\
        \bottomrule
    \end{tabular}
    \caption{\textbf{Comparison of representative CBVR benchmarks to SOVABench.} SOVABench provides a large number of queries, which are the samples used for evaluation. Retrieval performance is measured using mAP in most of the benchmarks.}
    \label{tab:retrieval_datasets}
    \vspace{-4pt}
\end{table*}

\subsection{Intra-pair Evaluation Protocol}
The intra-pair protocol evaluates the model’s ability to distinguish between visually similar but temporally inverse actions (\eg, open vs. close). Each opposite pair defines a binary retrieval set, where only the samples belonging to the opposite action act as non-relevant samples. The resulting metric, Pair-mAP, averages mAP over all opposite pairs:
\begin{equation}
    \text{Pair-mAP} = \frac{1}{|C|}\sum_{p=1}^{|C|} \text{mAP}_p
\end{equation}
where $C$ is the set of pairs and $\text{mAP}_p$ is the sample-level mAP obtained from the restricted set of samples belonging to the opposite action pair $p$.

All clips are self-contained and do not overlap with their opposite action. Since the aim of this protocol is to discriminate between pairs of opposite actions, no additional distracting samples are added. The intra-pair benchmark contains 2,300 queries covering 14 action classes (see Table~\ref{tab:opposing_actions} and Table \ref{tab:retrieval_datasets}). As in the inter-pair protocol, the number of queries is high, ensuring robust evaluation. The distribution of samples within each opposite action pair can be accessed in the supplementary material.

\subsection{Ethical Disclaim and License}
SOVABench is constructed from the MEVA and VIRAT surveillance datasets and inherits their ethical policies. Both source datasets were collected for research under controlled surveillance conditions, ensuring that no personally identifiable information is disclosed without consent. SOVABench does not redistribute or host original videos. Instead, it provides metadata and extraction procedures that allow reconstruction only upon obtaining access to MEVA and VIRAT under Creative Commons Attribution 4.0 (CC-BY-4.0) and VIRAT Video Dataset Usage Agreement, respectively. SOVABench's metadata is released under CC-BY-4.0. The authors disclaim any liability for annotation inaccuracies and for any unintended or inappropriate use of the SOVABench dataset.
\section{MLLM-to-Embedding Framework} \label{sec:method}
\begin{figure}[ht]
    \centering
    \includegraphics[width=0.8\linewidth]{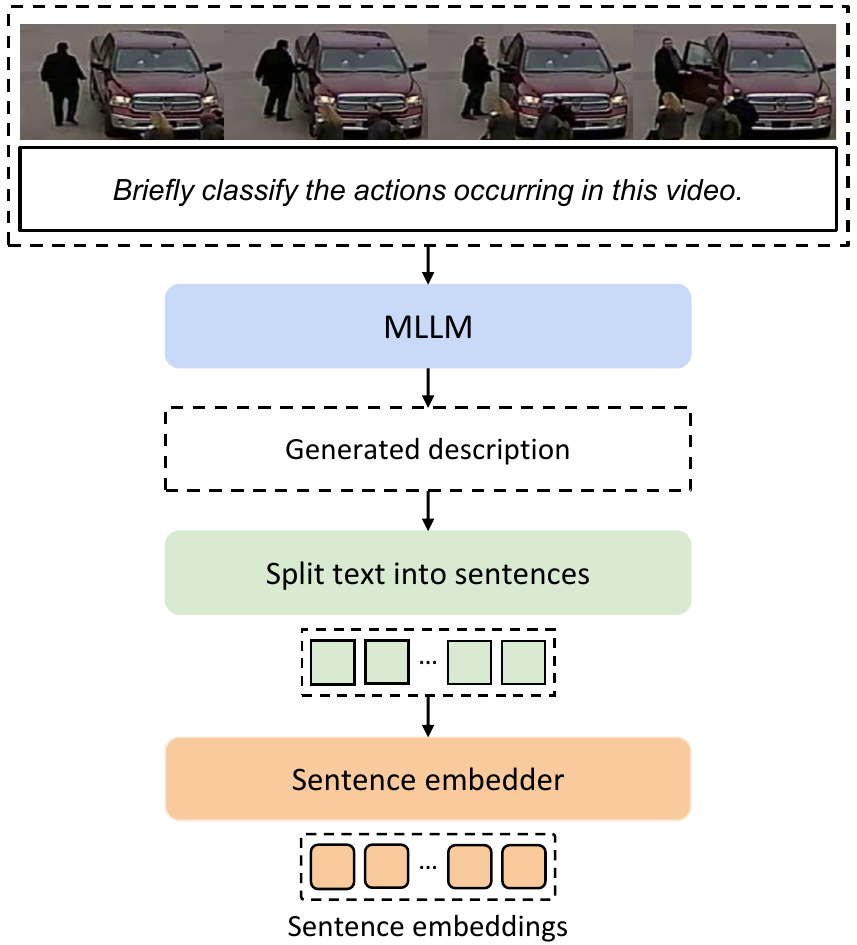}
    \caption{\textbf{Overview of the MLLM-to-Embedding framework.}
   Given an image/video and the textual instruction, an MLLM first generates a descriptive textual response. The output is then split into individual sentences, each encoded using a sentence-similarity text encoder. The similarity between two samples is computed using the maximum pairwise cosine similarity between their sentence embeddings (Equation \eqref{eq:similarity}). The filmstrip is extracted from SOVABench.}
    \label{fig:pipeline}
\end{figure}

Our goal is to obtain instruction-conditioned visual embeddings from an MLLM that can be used for classification and retrieval across different modalities, especially images and videos. We aim for a unified, training-free framework that leverages the reasoning capabilities of MLLMs. The overall pipeline is illustrated in Figure~\ref{fig:pipeline}.

Given a visual input $\mathcal{I}$ and a textual instruction $p$, we query an MLLM $\mathcal{G}$ to produce a textual response $t = \mathcal{G}(\mathcal{I}, p)$. This response captures the model’s interpretation of the visual input under the specified instruction. The output of the embedding pipeline is a variable-sized set of fixed-length vector representations $\mathbf{e}$. This set is constructed by first splitting the input text $t$ into sentences and then embedding each sentence independently using a sentence similarity encoder $\mathcal{E}$, as shown in Equation \eqref{eq:embeddings}.
\begin{equation} \label{eq:embeddings}
    \mathbf{e} = \bigcup_{t_i \in \mathrm{split}(t)} \mathcal{E}(t_i)
\end{equation}

The function $\mathrm{split}(t)$ is defined by first splitting the text $t$ into lines and then applying the NLTK sentence splitter \cite{bird2009nltk} to each line. Formally, $\mathrm{split}(t) = \{s \mid s \in \mathrm{sentence\_split}(l), l \in \mathrm{line\_split}(t)\}$.

To account for different levels of task specificity, we consider two prompting strategies:
\begin{enumerate}
    \item \textbf{General instruction:} The MLLM is prompted with a general instruction, namely ``Describe the image/video’’.
    \item \textbf{Task-aware instruction:} A prompt that specifies the type of information to extract (\eg, `List all pairwise spatial relations between objects’’ when the target dimension is spatial layout), thus directing the MLLM towards task-relevant semantics. The instructions are task-aware but do not include any information about class names or evaluation protocols. This ensures that the model is guided toward the intended dimension while preserving a zero-shot evaluation setting.
\end{enumerate}

To compute the similarity between two sets of embeddings $\mathbf{e}^{(1)}$ and $\mathbf{e}^{(2)}$, we use the pairwise similarity function $\mathcal{S}$:
\begin{equation} \label{eq:similarity}
    \mathcal{S}(\mathbf{e}^{(1)}, \mathbf{e}^{(2)}) = \max_{\mathbf{e}^{(1)}_i \in \mathbf{e}^{(1)}, \mathbf{e}^{(2)}_j \in \mathbf{e}^{(2)}}\mathrm{sim}(\mathbf{e}^{(1)}_i, \mathbf{e}^{(2)}_j),
\end{equation}
where $\mathrm{sim}(\cdot,\cdot)$ denotes cosine similarity. 

Several design choices address practical limitations of MLLM outputs. Splitting responses into sentences reduces sensitivity to irrelevant text in long-form descriptions. Using a maximum-similarity aggregator reflects scenarios where a single discriminative observation, such as a key spatial relation or motion cue, may suffice to determine similarity. An ablation study that evaluates the effect of incorporating sentence splitting with the maximum operator is shown in Section \ref{sec:split}.

The MLLM-to-Embedding framework relies on MLLMs to generate relevant visual semantics, to articulate these semantics as sentences, and to express temporal structure directly in their textual output. The embedding stage then treats the resulting sentences as black-box representations, focusing solely on encoding them for downstream classification and retrieval. In addition, task-aware prompting allows us to examine whether explicit instructions enhance the semantic quality of the generated embeddings.
\section{Experiments} \label{sec:experiments}
To evaluate the MLLM-to-Embedding framework, we use GTE-Large-8152 \cite{li2023towards, zhang2024mgte} as sentence encoder. The sensitivity to the choice of sentence encoder is low, as shown in ablation studies (see Section \ref{sec:sentence_emb}). For the MLLM’s text generation step, we employ greedy decoding to ensure deterministic outputs and reduce randomness. 

\subsection{Comparison against CLIP} \label{sec:image_comparison}
\begin{table*}[t]
\centering
\small
\setlength{\tabcolsep}{5pt}
\begin{tabular}{l | c c c c c c c | c c | c}
\toprule
\textbf{Model} & \multicolumn{2}{c}{\textbf{SpatialBench}} & \textbf{VSR} & \multicolumn{2}{c}{\textbf{What's Up}} & \textbf{CountBench} & \textbf{Visual7W-} & \textbf{Spatial Avg.} & \textbf{Count Avg.} & \textbf{Avg.} \\
 & \textbf{InD.} & \textbf{OutD.} & & \textbf{A} & \textbf{B} &  & \textbf{Count} & & & \\
\midrule
\textcolor{gray}{\textit{Random}} & \textcolor{gray}{29.8} & \textcolor{gray}{29.8} & \textcolor{gray}{50.0} & \textcolor{gray}{25.0} & \textcolor{gray}{25.0} & \textcolor{gray}{10.0} & \textcolor{gray}{10.0} & \textcolor{gray}{31.9} & \textcolor{gray}{17.5} & \textcolor{gray}{24.7} \\
CLIP-ViT-B/32 & 28.9 & 36.4 & 53.7 & 31.3 & 32.4 & 29.2 & 33.8 & 36.5 & 31.5 & 34.0 \\
CLIP-ViT-H-14\textsuperscript{\textdagger} \cite{liu2023vsr} & -- & -- & \textbf{54.5} & -- & -- & -- & -- & -- & -- & -- \\
XVLM 16M\textsuperscript{\textdagger} \cite{kamath2023s} & -- & -- & -- & 50.7 & 33.1  & -- & -- & -- & -- & -- \\
BLIP 14M\textsuperscript{\textdagger} \cite{kamath2023s} & -- & -- & -- & 38.8 & 38.2 & -- & -- & -- & -- & -- \\
Zhang et al.\textsuperscript{\textdagger} \cite{zhang2023zero} & -- & -- & -- & -- & -- & 30.7 & -- & -- & -- & -- \\
Singh et al.\textsuperscript{\textdagger} \cite{singh2024beyond} & -- & -- & -- & -- & -- & 34.2 & -- & -- & -- & -- \\
\midrule
\textbf{InternVL3.5 8B$_{\text{GENERAL}}$} & 37.9 & 35.9 & 51.6 & 63.3 & 27.9 & 43.9 & 50.4 & 43.3 & 47.2 & 45.3 \\
\textbf{MiniCPM-V 4.5$_{\text{GENERAL}}$} & 31.4 & 31.5 & 53.0 & \textbf{78.6} & 32.4 & 50.1 & 50.6 & 45.4 & 50.4 & 47.9 \\
\midrule
\textbf{InternVL3.5 8B$_{\text{TASK-AWARE}}$} & \textbf{38.6} & \textbf{37.7} & 52.0 & 64.8 & \textbf{46.3} & 68.6 & 52.5 & \textbf{47.9} & 60.6 & \textbf{54.3} \\
\textbf{MiniCPM-V 4.5$_{\text{TASK-AWARE}}$} & 35.0 & 34.6 & 51.4 & 36.7 & 36.3 & \textbf{76.4} & \textbf{55.0} & 38.8 & \textbf{65.7} & 52.3 \\
\bottomrule
\end{tabular}
\caption{\textbf{Comparison of CLIP and state-of-the-art approaches to the MLLM-to-Embedding framework across spatial understanding and object counting classification benchmarks.} For SpatialBench, we report the Indoor (InD.) and Outdoor (OutD.) datasets and for What's Up we report the subsets A and B. The final three columns average the results for spatial, counting, and spatial\&counting. In all cases, the metric is accuracy. \textsuperscript{\textdagger}Results obtained from the respective paper.}
\label{tab:image_results}
\end{table*}

Provided that our framework accepts both images and videos as input and can tackle classification and retrieval, we begin by evaluating its capabilities on image-based classification tasks that isolate specific aspects of scene understanding, namely spatial relations and object counting. These tasks provide a controlled setting in which the global embeddings of CLIP are known to be insufficient \cite{li2024exploring}, as they often fail to capture fine-grained relational or numerical semantics. Validating the framework under these constrained conditions allows us to assess whether instruction-conditioned, MLLM-derived embeddings offer a measurable advantage, before applying them to video retrieval.

For spatial understanding, we evaluate on SpatialBench \cite{wang2025spatialclip}, Visual Spatial Reasoning (VSR) \cite{liu2023vsr}, and What’s Up \cite{kamath2023s}. For object counting, we use CountBench \cite{paiss2023teaching} and Visual7W-Count, the latter derived from the counting questions in Visual7W \cite{zhu2016visual7w}. Declarative sentence choices for Visual7W-Count are produced automatically using ChatGPT and converting each annotated question and the original set of choices into CLIP-compatible sentences. Although perfect grammatical coherence across all generated sentences cannot be guaranteed, they only differ in numerical content, ensuring that any performance gaps reflect counting capability rather than linguistic artifacts.

As backbone MLLMs, we evaluate two state-of-the-art open-source models: InternVL-3.5 8B \cite{wang2025internvl3} and MiniCPM-V 4.5 \cite{yu2025minicpm}. Table \ref{tab:image_results} shows the results of these models against CLIP-ViT-B/32 \cite{radford2021learning} and state-of-the-art, train-free approaches. It is observed that, across all datasets except VSR, some configuration using the MLLM-to-Embedding framework achieves the highest classification performance. Both MLLMs exhibit substantial improvements over the CLIP baseline, with absolute gains of 11.5\% in spatial understanding and 34.2\% in object counting for the best configuration. The performance boost confirms the superiority of using MLLMs for these tasks.

When comparing General Instruction and Task-Aware Instruction prompting strategies, distinct trends emerge. For spatial understanding, task-awareness yields improvements for InternVL3.5 8B (+4.6\%) but reductions for MiniCPM-V 4.5 (-6.6\%). In contrast, task-awareness delivers pronounced boosts in object counting (+13.4\% for InternVL3.5 8B and +15.3\% for MiniCPM-V 4.5). We hypothesize that spatial understanding involves much more sub-tasks than object counting, for example, relative position layout, distance, size, and orientation. This wider range of sub-tasks leads to less useful instructions, as they do not guide the MLLM towards the specific sub-task but the general one. The prompts used for the task-aware configurations in each dataset are shown in the supplementary material.

\subsection{SOVABench}
\begin{table}[ht]
    \centering
    \begin{tabular}{ l | c c }
        \toprule
        \textbf{Model} & \textbf{Inter-} & \textbf{Intra-} \\
         & \textbf{pair} & \textbf{pair} \\
        \midrule
        \textcolor{gray}{Random} & \textcolor{gray}{3.4} & \textcolor{gray}{50.3} \\
        \midrule
        \textit{Contrastive Image-VLMs} & & \\
        CLIP-ViT-L-14 \cite{radford2021learning} & 29.1 & \underline{51.3} \\
        SigLIP2-Giant \cite{tschannen2025siglip} & \underline{30.6} & 51.1 \\
        MERU \cite{desai2023hyperbolic} & 28.6 & \underline{51.3} \\
        \midrule
        \textit{Contrastive Video-VLMs} & & \\
        VideoCLIP \cite{xu2021videoclip} & 34.5 & \underline{51.4} \\
        CLIP4Clip \cite{luo2022clip4clip} & \underline{36.6} & \underline{51.4} \\
        ActionCLIP \cite{wang2023actionclip} & 32.8 & 51.3 \\
        \midrule
        \textit{General MLLMs} & & \\
        \textbf{InternVL3.5 8B$_{\text{GENERAL}}$} & 27.7 & 51.7 \\
        \textbf{MiniCPM-V 4.5$_{\text{GENERAL}}$} & 34.4 & 52.5 \\
        \textbf{InternVL3.5 8B$_{\text{TASK-AWARE}}$} & 35.4 & 52.4 \\
        \textbf{MiniCPM-V 4.5$_{\text{TASK-AWARE}}$} & \underline{\textbf{38.3}} & \underline{53.6} \\
        \midrule
        \textit{Video-MLLMs} & & \\
        \textbf{VideoLLaVA 7B$_{\text{GENERAL}}$} \cite{zhang2024video} & 24.9 & 51.1 \\
        \textbf{VideoLlama3 7B$_{\text{GENERAL}}$} \cite{zhang2025videollama} & \underline{32.4} & 52.3 \\
        \textbf{VideoChat-R1 7B$_{\text{GENERAL}}$} \cite{li2025videochat} & 25.1 & 51.8 \\
        \textbf{VideoLLaVA 7B$_{\text{TASK-AWARE}}$} & 27.3 & 51.0 \\
        \textbf{VideoLlama3 7B$_{\text{TASK-AWARE}}$} & 29.2 & \underline{53.1} \\
        \textbf{VideoChat-R1 7B$_{\text{TASK-AWARE}}$} & 31.6 & 51.9 \\
        \midrule
        \textit{API MLLMs} & & \\
        \textbf{Gemini 2.5 Flash$_{\text{GENERAL}}$} & 27.9 & 52.6 \\
        \textbf{Qwen3-VL 235B A22B$_{\text{GENERAL}}$} \cite{yang2025qwen3} & 14.7 & 51.9 \\
        \textbf{Gemini 2.5 Flash$_{\text{TASK-AWARE}}$} & \underline{33.2} & \textbf{\underline{53.9}} \\
        \textbf{Qwen3-VL 235B A22B$_{\text{TASK-AWARE}}$} & 29.1 & 52.6 \\
        \bottomrule
    \end{tabular}
    \caption{\textbf{Performance comparison of models in SOVABench.} Models considered include contrastive image-VLMs, contrastive video-VLMs, general MLLMs, video-MLLMs, and API MLLMs. Metrics are mAP for SOVABench (Inter-pair) and Pair-mAP for SOVABench (Intra-pair).}
    \label{tab:sovabench_bench}
\end{table}

In the SOVABench evaluation, we benchmark a diverse set of models, including contrastive image-based VLMs\footnote{Video-level embeddings are obtained by averaging frame-level embeddings at a specified sampling rate.}, hyperbolic VLMs, and contrastive video-based VLMs. In addition, we use our framework to obtain embeddings from videos on general MLLMs that accept both images and videos, video-focused MLLMs, and API-based MLLMs. Frame sampling is set to 1 FPS where possible\footnote{ActionCLIP uses a fixed 32-frame sampling rate, while VideoCLIP operates at 30 FPS, according to their respective documentation.}, and sensitivity analysis indicates that performance is largely unaffected by this choice (see Section~\ref{sec:frame_rate}).

Table \ref{tab:sovabench_bench} summarizes the performance of the evaluated models. Under the inter-pair protocol, all models considerably outperform the random baseline (3.4 mAP), confirming that they exhibit some discriminative capability. Among all systems, the highest score is achieved by MiniCPM-V 4.5 using our framework and with task-aware prompting (38.3 mAP). At the category level, contrastive video-VLMs also show generally strong results, with CLIP4Clip obtaining the second best performance. 

In contrast, although Video-MLLMs are designed for video understanding, they do not show a consistent advantage compared to general-purpose MLLMs, indicating that their temporal modeling may not align with the short atomic actions in surveillance. In addition, the evaluation of API-based MLLMs, enabled by the fact that our framework treats MLLMs as black-box generators, reveals that even large proprietary models do not necessarily surpass open-source alternatives in this benchmark.

Regarding task specificity, task-aware prompting shows consistent gains, indicating that instruction design is a powerful lever for performance improvement. Combined with the reduced inference times of task-aware embeddings (see supplementary material), this points to a promising direction for further optimization.

For the intra-pair setting, all models perform only slightly above the random baseline (50.3 Pair-mAP), showing that evaluated systems struggle with this task. Although the MLLM-to-Embedding framework achieves better results, the proximity to random performance underscores the limitations of models in distinguishing the temporally inverse actions of SOVABench (Intra-pair). Previous work has also shown that multimodal models generally are poor at temporal-direction understanding \cite{liu2024tempcompass}, and SOVABench makes this weakness explicit.

To verify that the low scores arise from model shortcomings rather than from an ambiguous task, we conducted an error analysis, provided in the supplementary material. The results reveal that MLLMs often fail to generate accurate descriptions of actions that are straightforward to human observers. The dominant failure modes include: (1) generation errors and hallucinations, where the model invents objects or actions absent from the video; (2) under-description, where essential dynamic cues are omitted; and (3) temporal misunderstandings, where the action type is recognized but its temporal direction is reversed. These recurring failure modes also show that retrieval failures are primarily due to weaknesses in the generated descriptions rather than to the embedding stage alone.

\subsection{Ablation Study}
We perform an ablation study examining three key design factors in the proposed MLLM-to-Embedding framework: (1) the choice of sentence embedder, (2) the video frame sampling rate (only applies to SOVABench), and (3) the use of sentence splitting with the maximum aggregator. These components directly influence how textual descriptions are encoded, how temporal information is captured, and how multi-sentence outputs are compared against each other.

\subsubsection{Sentence Splitting with Maximum Aggregator} \label{sec:split}
\begin{table*}[t]
\centering
\small
\setlength{\tabcolsep}{5pt}
\begin{tabular}{l | c c c c c c c | c c | c}
\toprule
\textbf{Model} & \multicolumn{2}{c}{\textbf{SpatialBench}} & \textbf{VSR} & \multicolumn{2}{c}{\textbf{What's Up}} & \textbf{CountBench} & \textbf{Visual7W-} & \textbf{Spatial Avg.} & \textbf{Count Avg.} & \textbf{Avg.} \\
 & \textbf{InD.} & \textbf{OutD.} & & \textbf{A} & \textbf{B} &  & \textbf{Count} & & & \\
 \midrule
\textbf{InternVL3.5 2B$_{\text{GENERAL}}$} &  &  &  &  &  &  &  &  &  &  \\
\textbf{(w/ Sent. split + max)} & \underline{37.1} & \underline{39.0} & 51.9 & \textbf{\underline{62.4}} & \underline{28.7} & 39.1 & 50.4 & \textbf{\underline{43.8}} & 44.8 & \underline{44.3} \\
InternVL3.5 2B$_{\text{GENERAL}}$ &  &  &  &  &  &  &  &  &  &  \\
(w/o Sent. split + max) & 35.0 & 28.9 & \underline{53.0} & 32.5 & 23.0 & \underline{48.7} & \underline{50.7} & 34.5 & \underline{49.7} & 42.1 \\
\midrule
\textbf{InternVL3.5 2B$_{\text{TASK-AWARE}}$} &  &  &  &  &  &  &  &  &  &  \\
\textbf{(w/ Sent. split + max)} & \textbf{\underline{38.6}} & 32.1 & \textbf{\underline{54.2}} & \underline{45.6} & \textbf{\underline{48.0}} & 59.9 & 41.5 & \underline{43.7} & 50.7 & \textbf{\underline{47.2}} \\
InternVL3.5 2B$_{\text{TASK-AWARE}}$ &  &  &  &  &  &  &  &  &  &  \\
(w/o Sent. split + max) & 32.1 & \underline{32.7} & 51.4 & 31.3 & 29.2 & \textbf{\underline{61.3}} & \textbf{\underline{52.9}} & 35.3 & \textbf{\underline{57.1}} & 46.2 \\
\bottomrule
\end{tabular}
\caption{\textbf{Evaluation of the inclusion of the sentence splitting with maximum aggregator.} The model used is InternVL3.5 2B and both general and task-aware prompting strategies are reported. All metrics are accuracy.}
\label{tab:sentence_split}
\end{table*}

Table \ref{tab:sentence_split} shows the impact of introducing sentence splitting with maximum aggregator using InternVL3.5 2B\footnote{The 2-billion parameter version is used to streamline experiments.} in the image-based classification tasks from Section~\ref{sec:image_comparison}. The introduction of this module improves performance on spatial understanding benchmarks, where focusing on individual sentence-level descriptions and selecting the most informative one helps the model capture key spatial relationships more effectively. In contrast, this design choice shows a moderate reduction in counting objects. Nevertheless, overall average performance benefits from using sentence splitting with maximum aggregator and thus is found useful on the framework design. In addition, applying sentence splitting preserves semantics in arbitrarily long MLLM outputs.

\subsubsection{Sentence Embedder} \label{sec:sentence_emb}
\begin{table}[h]
    \centering
    \begin{tabular}{@{} l @{\hspace{3pt}} | @{\hspace{3pt}} c @{\hspace{5pt}} c @{\hspace{5pt}} c @{}}
        \textbf{Model} & \textbf{Params.} & \textbf{Emb. size} & \textbf{mAP} \\
        \midrule
        CLIP-ViT-L-14 (Text tower) & 123M & 768 & 33.5 \\
        all-MiniLM-L6-v2\footnotemark & 23M & 384 & 37.8 \\
        EmbeddingGemma \cite{vera2025embeddinggemma} & 308M & 768 & 36.7 \\
        GTE-Large-8152 & 409M & 1024 & \textbf{38.3} \\
        mE5 Large Instruct \cite{wang2024multilingual} & 560M & 1024 & 37.7 \\
        Qwen3 Embedding 0.6B \cite{zhang2025qwen3} & 596M & 1024 & 36.9 \\
    \end{tabular}
    \caption{\textbf{Sensitivity of the MLLM-to-Embedding framework of the choice of sentence embedder.} Executions are done with MiniCPM-V 4.5$_{\text{TASK-AWARE}}$ on SOVABench (Inter-pair) with 1 FPS. The number of parameters (Params.) and the dimensionality of the embedding space (Emb. size) are reported.}
    \label{tab:sentence_emb}
\end{table}

We compare several state-of-the-art text embedding models to assess whether the choice of sentence embedder influences retrieval performance. The comparison is conducted on SOVABench (Inter-pair) using the MiniCPM-V 4.5$_{\text{TASK-AWARE}}$ configuration, which provides the strongest results among our evaluated settings. As shown in Table~\ref{tab:sentence_emb}, only minor variations are observed between embedders, indicating that the framework is largely robust to this component. Moreover, neither the number of parameters nor the dimensionality of the embedding space shows a clear relationship with performance. Given its slightly superior results, we adopt GTE-Large-8152 as the default embedder.

\footnotetext{https://huggingface.co/sentence-transformers/all-MiniLM-L6-v2}

\subsubsection{Frame Sampling Rate} \label{sec:frame_rate}
To evaluate the role of temporal resolution, we vary the frame sampling rate when processing videos with MiniCPM-V 4.5$_{\text{TASK-AWARE}}$ in SOVABench (both protocols). As shown in Table \ref{tab:fps}, increasing the sampling density does not yield improvements in SOVABench, suggesting that representative frames are sufficient to capture action information in this benchmark. As a result, we use 1 FPS for efficiency without loss of performance.

\begin{table}[ht]
    \centering
    \begin{tabular}{l | c c}
        \textbf{Frame rate} & \textbf{Inter-Pair} & \textbf{Intra-Pair} \\
         & \textbf{(mAP)} & \textbf{(Pair-mAP)} \\
         \midrule
         1 FPS & \textbf{38.3} & 53.6 \\
         3 FPS & 37.3 & \textbf{54.0} \\
         5 FPS & 36.3 & 53.8 \\
         7 FPS & 36.2 & 53.4 \\
    \end{tabular}
    \caption{\textbf{Sensitivity of SOVABench evaluations on the sampling rate.} Executions are done with MiniCPM-V 4.5$_{\text{TASK-AWARE}}$.}
    \label{tab:fps}
\end{table}

\section{Conclusions and Future Work} \label{sec:conclusions}
We introduced SOVABench, a challenging benchmark for action retrieval in vehicle-related surveillance scenarios. Its two complementary protocols (inter-pair and intra-pair retrieval) jointly provide both a global assessment of action-level representation quality and a measure of temporal direction understanding, enabling analysis of failure modes in action description. In addition, we construct the MLLM-to-Embedding framework to obtain sentence-level embeddings from MLLMs, enabling both retrieval and classification. Our experiments demonstrate that even a simple, instruction-following framework improves performance compared to contrastive methods, while providing interpretable representations.

We expect SOVABench to facilitate further research in action retrieval for vehicle-surveillance scenarios. Future work should explore the improvement of prompting strategies and embedding mechanisms of MLLMs to encode temporal progression and action dynamics more precisely.

\section*{Acknowledgments}
This work has been partially supported by the Spanish project PID2022-136436NB-I00, by ICREA under the ICREA Academia programme, and by the Milestone Research Program at the University of Barcelona.

{
    \small
    \bibliographystyle{ieeenat_fullname}
    \bibliography{main}
}

\clearpage
\appendix
\renewcommand{\thefigure}{A.\arabic{figure}}
\setcounter{figure}{0}
\begin{figure*}[ht]
  \centering
  \begin{subfigure}{0.23\textwidth}
    \centering
    \begin{tikzpicture}
      \pie[radius=1, sum=auto]{303/Open, 301/Close}
    \end{tikzpicture}
    \caption{Open / Close vehicle door}
  \end{subfigure}\hfill
  \begin{subfigure}{0.23\textwidth}
    \centering
    \begin{tikzpicture}
      \pie[radius=1, sum=auto]{261/Enter, 212/Exit}
    \end{tikzpicture}
    \caption{Enter / Exit vehicle}
  \end{subfigure}\hfill
  \begin{subfigure}{0.23\textwidth}
    \centering
    \begin{tikzpicture}
      \pie[radius=1, sum=auto]{252/Left, 204/Right}
    \end{tikzpicture}
    \caption{Turn left / right}
  \end{subfigure}\hfill
  \begin{subfigure}{0.23\textwidth}
    \centering
    \begin{tikzpicture}
      \pie[radius=1, sum=auto]{167/Start, 215/Stop}
    \end{tikzpicture}
    \caption{Start / Stop}
  \end{subfigure}
  \vspace{8pt}
  \begin{subfigure}{0.32\textwidth}
    \centering
    \begin{tikzpicture}
      \pie[radius=1, sum=auto]{83/Drive forward, 90/Reverse}
    \end{tikzpicture}
    \caption{Drive forward / Reverse}
  \end{subfigure}\hfill
  \begin{subfigure}{0.32\textwidth}
    \centering
    \begin{tikzpicture}
      \pie[radius=1, sum=auto]{54/Load, 61/Unload}
    \end{tikzpicture}
    \caption{Load / Unload vehicle}
  \end{subfigure}\hfill
  \begin{subfigure}{0.32\textwidth}
    \centering
    \begin{tikzpicture}
      \pie[radius=1, sum=auto]{49/Open, 48/Close}
    \end{tikzpicture}
    \caption{Open / Close trunk}
  \end{subfigure}
  \caption{\textbf{Number of video samples per action class within each opposite pair in the SOVABench (Intra-Pair) benchmark.}}
  \label{fig:intraclass}
\end{figure*}
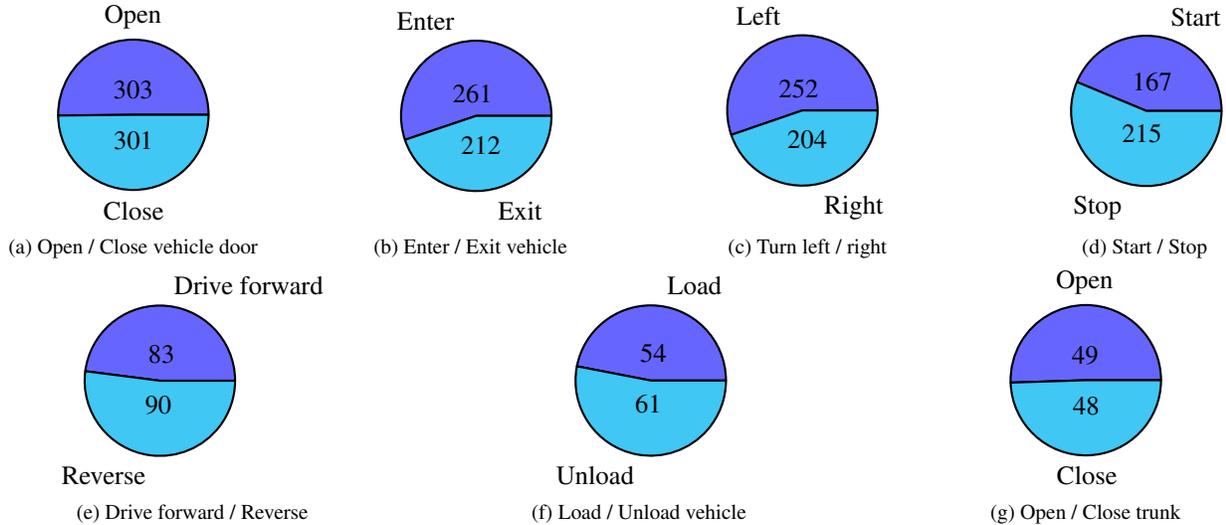

The supplementary material contains the following information.
\begin{itemize}
    \item Section~\ref{sec:intra_distribution} shows the sample numbers of the SOVABench (Intra-pair) classes.
    \item Section~\ref{sec:instructions} shows the instructions used in each dataset for the task-aware prompting strategy.
    \item Section~\ref{sec:queries} shows the effect of not incorporating distracting samples into SOVABench (Inter-pair).
    \item Section~\ref{sec:efficiency} analyzes the latency differences between models.
    \item Section~\ref{sec:error} performs an error analysis on some SOVABench (Intra-pair) samples.
\end{itemize}

\section{Intra-pair Distribution} \label{sec:intra_distribution}
Figure~\ref{fig:intraclass} shows the number of video clips per action class within the set of opposite action pairs in SOVABench (Intra-pair). All pairs have close to even distributions.

\section{Instruction per Dataset} \label{sec:instructions}
These are the instructions provided to MLLMs in each dataset:
\begin{itemize}
    \item \textbf{SpatialBench:} List all spatial relationships between objects (e.g., position, size, distance, or orientation) in short sentences.
    \item \textbf{VSR:} List all pairwise spatial relations between objects in the image.
    \item \textbf{What's Up:} List all pairwise spatial relations between objects in the image.
    \item \textbf{CountBench:} Describe the image in a short caption that accurately states the number of main objects (in words) and includes a brief descriptive phrase.
    \item \textbf{Visual7W-Count:} Describe the image in a short caption that accurately states the number of main objects (in words) and includes a brief descriptive phrase.
    \item \textbf{SOVABench:} Briefly classify the actions occurring in this video. (+ System prompt: You are an expert video analysis model specialized in action recognition. Focus on how subjects and objects change and move over time rather than on static appearances or backgrounds. Infer the actions by reasoning about motion, temporal progression, and interactions across the video frames.)
\end{itemize}

\section{Inter-pair Evaluation Constrained to Queries} \label{sec:queries}
Table \ref{tab:sovabench_bench_queries} shows the performance of the models in the inter-pair evaluation protocol constrained to the set of queries (1,423 samples). The results indicate similar trends to those found in the default inter-pair evaluation setting. However, this constrained evaluation reveals a larger advantage for MLLMs with MLLM-to-Embedding framework over contrastive VLMs: more MLLM-based configurations surpass the strongest contrastive baseline, and the performance gaps widen (best MLLM: 44.8 mAP, best contrastive VLM: 41.4 mAP). This setting is also more challenging than the default one despite returning higher absolute values, since the margin from random performance (23.7 mAP) is compressed. This increased difficulty arises from the higher similarity among samples, where all clips depict vehicle-related activities and the distracting human-only clips are removed. In summary, when samples are more similar to each other (without distractors), the advantage of MLLMs becomes more pronounced.

\renewcommand{\thetable}{C.\arabic{table}}
\setcounter{table}{0}
\begin{table}[ht]
    \centering
    \begin{tabular}{@{} l @{\hspace{3pt}} | @{\hspace{3pt}} c @{\hspace{3pt}} | @{\hspace{3pt}} c @{}}
        \toprule
        \textbf{Model} & \textbf{Efficiency} & \textbf{Inter-pair} \\
        &  & \textbf{(Constr.)} \\
        \midrule
        \textcolor{gray}{Random} & \textcolor{gray}{--} & \textcolor{gray}{23.7} \\
        \midrule
        \textit{Contrastive Image-VLMs} & & \\
        CLIP-ViT-L-14 & \textbf{\underline{22.88}} & 37.4 \\
        SigLIP2-Giant & 3.94 & \underline{38.7}  \\
        MERU & 21.25 & 36.9 \\
        \midrule
        \textit{Contrastive Video-VLMs} & \\
        VideoCLIP & 0.47 & \underline{41.4} \\
        CLIP4Clip & \underline{9.63} & 36.6 \\
        ActionCLIP & 7.78 & 36.0 \\
        \midrule
        \textit{General MLLMs} & \\
        \textbf{InternVL3.5 8B$_{\text{GENERAL}}$} & 0.26 & 39.4 \\
        \textbf{MiniCPM-V 4.5$_{\text{GENERAL}}$} & 0.10 & 42.2 \\
        \textbf{InternVL3.5 8B$_{\text{TASK-AWARE}}$} & \underline{0.33} & 44.2 \\
        \textbf{MiniCPM-V 4.5$_{\text{TASK-AWARE}}$} & 0.26 & \underline{\textbf{44.8}} \\
        \midrule
        \textit{Video-MLLMs} & & \\
        \textbf{VideoLLaVA 7B$_{\text{GENERAL}}$} & 0.16 & 33.1 \\
        \textbf{VideoLlama3 7B$_{\text{GENERAL}}$} & 0.42 & 40.7 \\
        \textbf{VideoChat-R1 7B$_{\text{GENERAL}}$} & 0.06 & 36.3 \\
        \textbf{VideoLLaVA 7B$_{\text{TASK-AWARE}}$} & 0.22 & 35.8 \\
        \textbf{VideoLlama3 7B$_{\text{TASK-AWARE}}$} & \underline{0.44} & 40.5 \\
        \textbf{VideoChat-R1 7B$_{\text{TASK-AWARE}}$} & 0.13 & \underline{42.7} \\
        \midrule
        \textit{API MLLMs} & & \\
        \textbf{Gemini 2.5 Flash$_{\text{GENERAL}}$} & -- & 38.1 \\
        \textbf{Qwen3-VL 235B A22B$_{\text{GENERAL}}$} & -- & 31.6 \\
        \textbf{Gemini 2.5 Flash$_{\text{TASK-AWARE}}$} & -- & \underline{43.0} \\
        \textbf{Qwen3-VL 235B A22B$_{\text{TASK-AWARE}}$} & -- & 42.8 \\
        \bottomrule
    \end{tabular}
    \caption{\textbf{Performance comparison of models in SOVABench (Inter-pair) restricted to the set of query samples, and efficiency comparison.} Efficiency is calculated as instances processed per second in the evaluation of the default SOVABench (Inter-pair) benchmark including distracting samples. SOVABench (Inter-pair) uses mAP.}
    \label{tab:sovabench_bench_queries}
\end{table}

\section{Efficiency Analysis} \label{sec:efficiency}
Table \ref{tab:sovabench_bench_queries} shows the number of instances processed per second for each model\footnote{All values are obtained using GPUs of type NVIDIA GeForce RTX 3090.}. The table confirms that MLLMs are naturally heavier and, therefore, slower than usual contrastive VLMs. However, the key observation is that task-aware configurations consistently deliver faster inference than their general counterparts. This means that task-aware prompting offers not only performance gains, but also an efficiency advantage, allowing models to produce shorter and more task-relevant outputs.
\section{Error Analysis of SOVABench (Intra-pair)} \label{sec:error}
We analyze the answers generated by MiniCPM-V 4.5$_{\text{TASK-AWARE}}$ in the opposite action pair $<$Open trunk, Close trunk$>$ in the intra-pair evaluation protocol. Based on the error analysis, we group the errors detected into a set of 4 error modes. The subsequent list provides the description of each error mode and its counts from the analysis (shown in brackets). The list is prioritized by severity, meaning that if an error belongs to more than one mode, the most serious one is assigned. Examples of each error mode are illustrated in Figure~\ref{fig:movie_strip_multi}.

\begin{enumerate}
    \item \textbf{Generation Errors / Hallucination (34):} The model produces semantically incorrect descriptions or fabricated entities, objects or actions.
    \item \textbf{Under-Description (12):} The model provides a description that is factually correct but insufficiently expressive, omitting dynamic event details (\eg, omitting any reference to the trunk of a vehicle when there is a trunk action).
    \item \textbf{Temporal Misunderstanding (16):} The model identifies the correct action type but assigns an incorrect temporal direction, such as reversing ``opening'' and ``closing'' a trunk.
    \item \textbf{Action Assumption (9):} The model correctly detects the high-level action but omits required intermediate steps, resulting in an incomplete or overly compressed event sequence (\eg, describing ``walking to a vehicle and loading an object'' without mentioning the intermediate action of opening the trunk). Although not an error, it complicates the retrieval of actions belonging to the same class.
\end{enumerate}

Following the numbers for each error mode, there are only 26 totally correct samples out of 97. In addition, the examples in Figure~\ref{fig:movie_strip_multi} show that most retrieval failures originate from textual-generation issues rather than embedding quality alone.

\renewcommand{\thefigure}{E.\arabic{figure}}
\setcounter{figure}{0}
\begin{figure*}[ht]
    \centering
    \captionsetup[subfigure]{skip=1pt}
    \begin{subfigure}{\textwidth}
        \begin{tcolorbox}
            \begin{minipage}{\linewidth}
                \centering
                \includegraphics[width=0.5\linewidth]{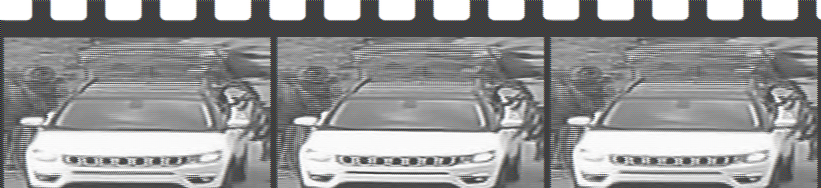}
            \end{minipage}
            \\
            \textbf{Model Answer:} A person is standing on the left side of the car, and another person is standing on the right side of the car. The person on the right side of the car opens the trunk of the car, and the person on the left side of the car also opens the trunk of the car. The person on the right side of the car is taking something out of the trunk of the car, and the person on the left side of the car is also taking something out of the trunk of the car. The person on the right side of the car is taking something out of the trunk of the car and putting it into a bag, and the person on the left side of the car is also taking something out of the trunk of the car and putting it into a bag. ... \\
            \textbf{Observation:} Degenerated generation.
        \end{tcolorbox}
        \caption{Generation Errors / Hallucination}
    \end{subfigure}
        
    \begin{subfigure}{\textwidth}
        \begin{tcolorbox}
            \begin{minipage}{\linewidth}
                \centering
                \includegraphics[width=0.5\linewidth]{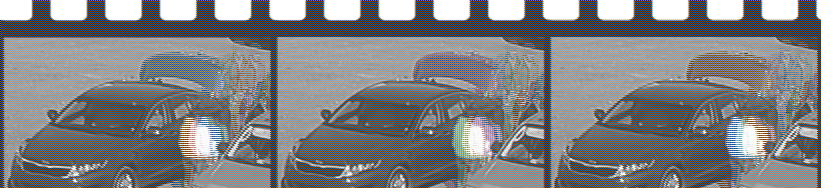}
            \end{minipage}
            \\
            \textbf{Model Answer:} A person is opening the trunk of a car, and then the trunk is closed. \\
            \textbf{Observation:} Incoherent answer.
        \end{tcolorbox}
        \caption{Generation Errors / Hallucination}
    \end{subfigure}
    \begin{subfigure}{\textwidth}
        \begin{tcolorbox}
            \begin{minipage}{\linewidth}
                \centering
                \includegraphics[width=0.5\linewidth]{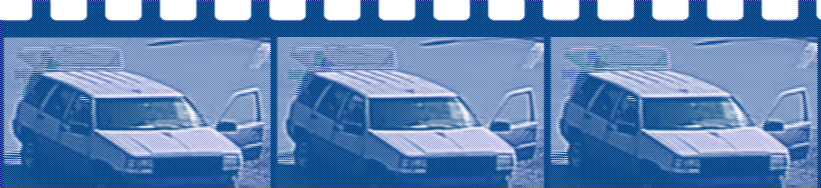}
            \end{minipage}
            \\
            \textbf{Model Answer:} A person is standing on the roof of a car, the car door is open, the car trunk is open, and the person is moving. \\
            \textbf{Observation:} Hallucinated illogical person standing on the roof of a car.
        \end{tcolorbox}
        \caption{Generation Errors / Hallucination}
    \end{subfigure}
    \begin{subfigure}{\textwidth}
        \begin{tcolorbox}
            \begin{minipage}{\linewidth}
                \centering
                \includegraphics[width=0.5\linewidth]{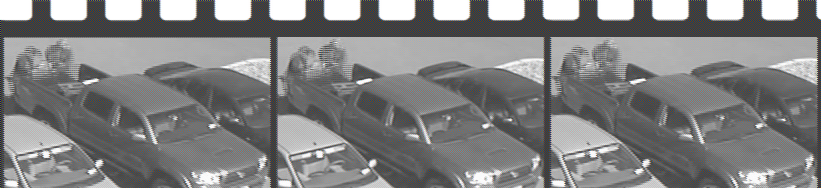}
            \end{minipage}
            \\
            \textbf{Model Answer:} Two people are standing in the back of a pickup truck, and they are looking down at something. \\
            \textbf{Observation:} Statement with no key actions.
        \end{tcolorbox}
        \caption{Under-Description}
    \end{subfigure}
    \begin{subfigure}{\textwidth}
        \begin{tcolorbox}
            \begin{minipage}{\linewidth}
                \centering
                \includegraphics[width=0.5\linewidth]{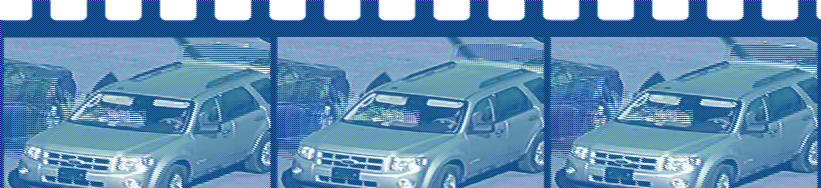}
            \end{minipage}
            \\
            \textbf{Model Answer:} A car is parked, and a person is getting in. \\
            \textbf{Observation:} Trunk-related actions are not mentioned.
        \end{tcolorbox}
        \caption{Under-Description}
    \end{subfigure}
\end{figure*}

\begin{figure*}[ht]\ContinuedFloat
    \centering
    \captionsetup[subfigure]{skip=1pt}
    \begin{subfigure}{\textwidth}
        \begin{tcolorbox}
            \begin{minipage}{\linewidth}
                \centering
                \includegraphics[width=0.5\linewidth]{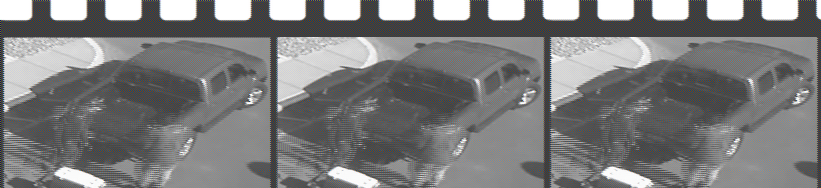}
            \end{minipage}
            \\
            \textbf{Model Answer:} A man is pushing a cooler on a dolly towards a blue truck. Another man is standing next to the truck, opening the tailgate. The man with the cooler moves the dolly towards the truck, and the other man continues to hold the tailgate open. \\
            \textbf{Observation:} Wrong temporal understanding since the tailgate is being closed.
        \end{tcolorbox}
        \caption{Temporal Misunderstanding}
    \end{subfigure}
    \begin{subfigure}{\textwidth}
        \begin{tcolorbox}
            \begin{minipage}{\linewidth}
                \centering
                \includegraphics[width=0.5\linewidth]{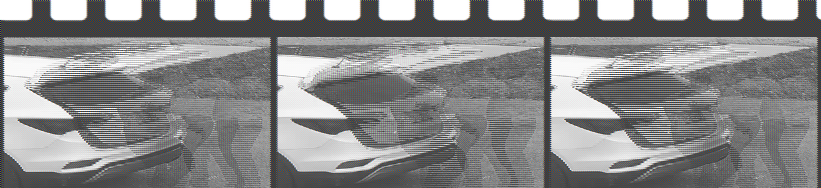}
            \end{minipage}
            \\
            \textbf{Model Answer:} A man walks to the back of a car, opens the trunk, and places a box inside. \\
            \textbf{Observation:} Although the temporal direction within the close / open trunk pair is correctly determined, the next action should be unloading a box, not loading.
        \end{tcolorbox}
        \caption{Temporal Misunderstanding}
    \end{subfigure}
    \begin{subfigure}{\textwidth}
        \begin{tcolorbox}
            \begin{minipage}{\linewidth}
                \centering
                \includegraphics[width=0.5\linewidth]{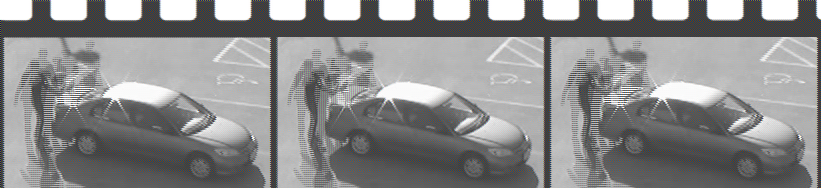}
            \end{minipage}
            \\
            \textbf{Model Answer:} A man is walking towards a car, and another man is standing next to the car. The man in the black shirt walks up to the car and the man in the red shirt puts a box in the trunk. \\
            \textbf{Observation:} No mention to opening the trunk as intermediate action.
        \end{tcolorbox}
        \caption{Action Assumption}
    \end{subfigure}
    \begin{subfigure}{\textwidth}
        \begin{tcolorbox}
            \begin{minipage}{\linewidth}
                \centering
                \includegraphics[width=0.5\linewidth]{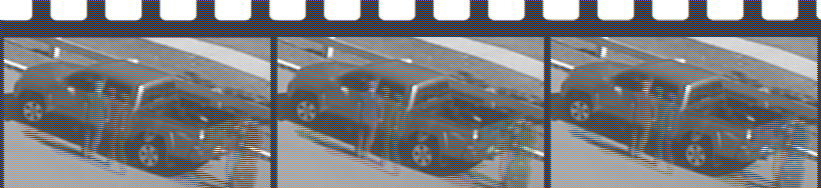}
            \end{minipage}
            \\
            \textbf{Model Answer:} A man is walking to the back of a car, a man is standing in the back of a car, a man is loading a suitcase into the back of a car. \\
            \textbf{Observation:} No mention to opening the tailgate as intermediate action.
        \end{tcolorbox}
        \caption{Action Assumption}
    \end{subfigure}
    \begin{subfigure}{\textwidth}
        \begin{tcolorbox}
            \begin{minipage}{\linewidth}
                \centering
                \includegraphics[width=0.5\linewidth]{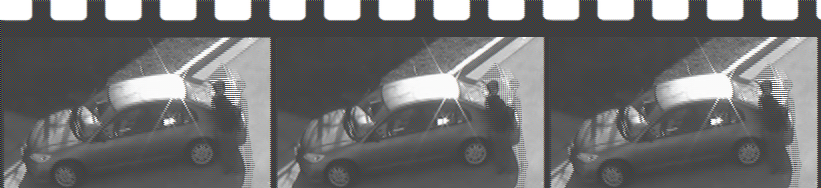}
            \end{minipage}
            \\
            \textbf{Model Answer:} A man is standing next to a car, then he opens the trunk of the car.
        \end{tcolorbox}
        \caption{Correct}
    \end{subfigure}
    \vspace{-2em}
    \caption{\textbf{Examples of generation errors obtained from MiniCPM-V 4.5$_\text{TASK-AWARE}$ in the $<$Open trunk, Close trunk$>$ pair.} Errors are prioritized from more to less sever, including Generation Errors / Hallucination (a)-(c), Under-Description (d)-(e), Temporal Misunderstanding (f)-(g), and Action Assumption (h)-(i). In addition, (j) shows a successful case. Each example is composed by a filmstrip of the video, the model answer and an observation indicating why is wrong.}
    \label{fig:movie_strip_multi}
\end{figure*}

\end{document}